\documentclass[runningheads]{llncs}
\usepackage[T1]{fontenc}
\usepackage{amsmath,amssymb}
\usepackage{graphicx}
\usepackage[colorlinks=true,citecolor=black,linkcolor=black,urlcolor=blue]{hyperref}
\usepackage{booktabs, caption}
\usepackage{placeins}

\usepackage{tikz,graphics,color,float,epsf,caption,subcaption}
\usepackage{svg}
\usepackage{graphicx}
\begin{document}
\title{AI Age Discrepancy: A Novel Parameter for Frailty Assessment in Kidney Tumor Patients}
\titlerunning{AI Age Discrepancy}
%
\author{
Rikhil Seshadri\inst{1}\thanks{These authors contributed equally.} \and
Jayant Siva\inst{1}$^*$ \and
Angelica Bartholomew\inst{1} \and
Clara Goebel\inst{1} \and
Gabriel Wallerstein-King\inst{1} \and
Beatriz López Morato\inst{1} \and
Nicholas Heller\inst{1} \and
Jason Scovell\inst{1} \and
Rebecca Campbell\inst{1} \and
Andrew Wood\inst{1} \and
Michal Ozery-Flato\inst{2} \and
Vesna Barros\inst{2} \and
Maria Gabrani\inst{3} \and
Michal Rosen-Zvi\inst{2} \and
Resha Tejpaul\inst{4} \and
Vidhyalakshmi Ramesh\inst{4} \and
Nikolaos Papanikolopoulos\inst{4} \and
Subodh Regmi\inst{4} \and
Ryan Ward\inst{1} \and
Robert Abouassaly\inst{1} \and
Steven C. Campbell\inst{1} \and
Erick Remer\inst{1} \and
Christopher Weight\inst{1}
}

\authorrunning{
Rikhil Seshadri and Jayant Siva et al.
}
%
\institute{
Cleveland Clinic, Cleveland, United States \and
IBM Research, Haifa, Israel \and
IBM Research, Zurich, Switzerland \and
University of Minnesota, Minneapolis, United States
}
\maketitle              
\begin{abstract}
Kidney cancer is a global health concern, and accurate assessment of patient frailty is crucial for optimizing surgical outcomes. This paper introduces AI Age Discrepancy, a novel metric derived from machine learning analysis of preoperative abdominal CT scans, as a potential indicator of frailty and postoperative risk in kidney cancer patients. This retrospective study of 599 patients from the 2023 Kidney Tumor Segmentation (KiTS) challenge dataset found that a higher AI Age Discrepancy is significantly associated with longer hospital stays and lower overall survival rates, independent of established factors. This suggests that AI Age Discrepancy may provide valuable insights into patient frailty and could thus inform clinical decision-making in kidney cancer treatment.

\keywords{Kidney Cancer \and Machine Learning \and Frailty.}
\end{abstract}
\section{Introduction}
Kidney cancer is a malignancy that originates from the tissues of the kidney or renal pelvis and is primarily discovered through the presence of a kidney tumor on abdominal imaging. While some kidney tumors are benign and do not carry a risk for metastasis, most are malignant \cite{Chawla2006}. In 2020, kidney cancer made up 2.2\% of all cancer diagnoses and 1.8\% of all cancer-related fatalities globally. Kidney cancer poses a substantial global health burden with an age-standardized incidence rate of 6.1 per 100,000 in men \cite{Sung2021}.

\begin{figure}[H]
    \centering
    \includegraphics[width=1\linewidth]{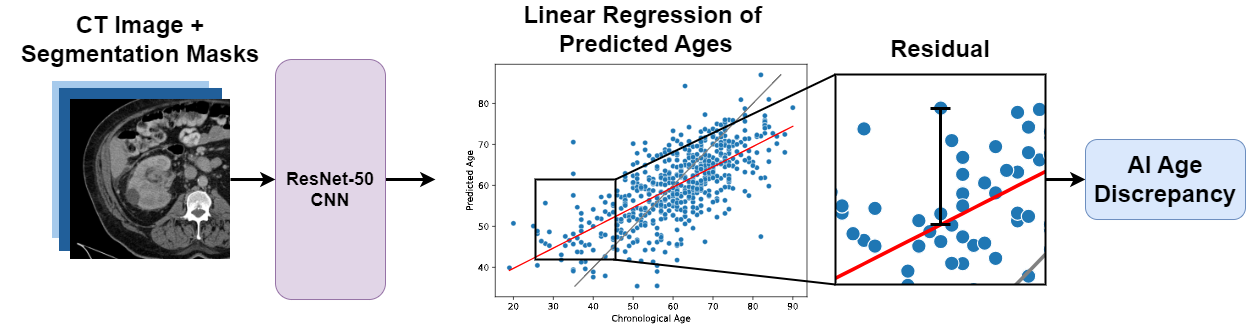}
    \caption{\textbf{Proposed AI Age Discrepancy Process.} ResNet-50 convolutional neural networks are trained to predict age from preoperative CT images. The linear regression line of predictions is calculated. Residuals from predicated age to regression line are taken and then normalized to obtain the AI Age Discrepancy.}
\end{figure}
\vspace{-0.6cm}
Once a kidney tumor has been identified on imaging, current treatments involve various surgical approaches, each with distinct advantages and challenges. Open nephrectomy, which uses an incision to remove the kidney in its entirety, provides direct access into the retroperitoneum but results in longer recovery times and higher complication risks \cite{Kunath2017}. In contrast, minimally invasive nephrectomy, including robotic-assisted techniques, uses small incisions and a camera, often resulting in shorter hospital stays, reduced pain, and faster recovery. However, these minimally invasive procedures require specialized skills and may not be as widely available as other interventions \cite{Kunath2017}. Cryoablation is another minimally invasive procedure used for small, early-stage renal tumors. This technique involves inserting a probe into the tumor to freeze and destroy cancer cells under CT imaging guidance. Cryoablation is best suited to patients who cannot undergo surgery, as it carries a higher risk of local recurrence when compared to surgical options \cite{Bisbee2022}. Nephron-sparing procedures, including partial nephrectomy and cryoablation, are often preferred over radical nephrectomy for smaller renal tumors because they better preserve long-term renal function and lower non-cancer-related mortality rates. Despite being the current gold standard, partial nephrectomy is technically more challenging and associated with a higher rate of minor complications compared to radical nephrectomy \cite{Kalogirou2017}.

Tumor stage, grade, and surgical approach are well-established factors influencing both length of hospital stay and survival after kidney tumor surgery \cite{Campbell2021}\cite{Kunath2017}. The AUA guidelines delineate tumor stage, grade, and histology as primary determinants of post-intervention prognosis \cite{Campbell2021}. Upstaging during surgery (e.g., cT1 to pT3a) indicates a poorer prognosis, highlighting the importance of accurate preoperative staging \cite{Veccia2020}. Surgical approaches like laparoscopy and robotic surgery also offer faster recovery compared to open nephrectomy \cite{Kunath2017}. Choosing the appropriate surgical approach depends on tumor size, surgeon expertise, and, the overall health status or 'frailty' of the patient in question. These factors can play a significant role in patients' length of hospital stay and overall survival. 


Previous research has found that deep neural networks can be trained to predict a patient’s age from CT scans. For instance, Kerber et al. (2023) proposed a deep learning model for age estimation from thoracic and abdominal CT scans, emphasizing the importance of prediction reliability and identifying regions significant for age estimation \cite{Kerber2023}. Azarfar et al. (2023) also developed a model for age estimation from chest CT scans, suggesting that CT-estimated age may be a useful addition to the calculation of cancer risk \cite{Azarfar2023}. These prior works fall into the emerging category of ``biological age'' quantification \cite{babyn2023ai} which aims to use biological measurements to quantify age-related changes in a patient's health. 

The present study aims to build on these findings by considering AI Age Discrepancy as a predictor of frailty in the context of kidney tumor treatment. This novel parameter may provide a valuable independent signal about patients' overall health status to complement established frailty indices such as the Charlson Comorbidity Index (CCI) \cite{Sundararajan102004}.


\section{Methods}

\subsection{Dataset}
This retrospective study included 599 patients who underwent treatment for suspected renal malignancy at a single institution from 2010 to 2021. The patients were taken from the publicly available 2023 Kidney Tumor Segmentation (KiTS) challenge held in conjunction with the International Conference on Medical Image Computing and Computer-Assisted Interventions (MICCAI) in 2023 \cite{Heller2023}. 9 patients under the age of 18 were excluded since the mechanisms for age-related frailty are not likely to impact adolescents. All patients underwent preoperative contrast-enhanced abdominal CT imaging in either the arterial or venous phase. The original compilation and public release of the KiTS23 cohort received approval from the University of Minnesota Institutional Review Board. This subsequent analysis of de-identified data was deemed exempt from IRB review. 

\subsection{Age Prediction Model}
A ResNet-50 architecture with pre-trained weights from the ImageNet database was fine-tuned to predict age as a continuous variable. 2-dimensional views were extracted from all three anatomical planes, and three channels were constructed containing (1) the CT image Hounsfield Unit attenuation values, (2) a mask representing the tumor segmentation, and (3) a mask representing the affected kidney. Scans in the training set were randomly sampled for training, and 2D views were sampled from selected scans with weighted probabilities corresponding to the fraction of tumor voxels in each view. Final age predictions were obtained by computing a weighted average of the predictions made on all views, with weight again determined by fraction of tumor voxels.  

\begin{figure}[H]
    \centering
    \includegraphics[width=0.7\linewidth]{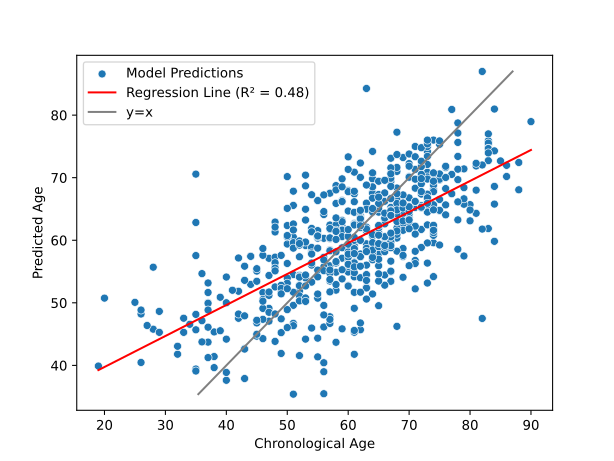}
    \caption{\textbf{Scatter plot of patient predicted ages (blue), linear regression line of predicted ages (red), and line of perfect predictions (grey).} AI Age Discrepancy is the normalized residual from the predicted age to the linear regression line.}
    \label{fig:single_svg}
\end{figure}

5-fold cross-validation was repeated three times to obtain average test-set age predictions for the full dataset. A least squares regression fit between the predicted age and the chronological age of the patient is shown in \textit{Figure \ref{fig:single_svg}}. Some regression to the mean is observed, and must be adjusted for in downstream analyses. Therefore, the residuals between the predicted age and the regression line (as opposed to y=x) were calculated, and the residuals were then normalized with a mean of 0 and a standard deviation of 1. We refer to these normalized residuals as the patients' AI Age Discrepancy. We hypothesize that the AI Age Discrepancy captures a measure of a patient's biological health, potentially reflecting underlying health status that may not be fully revealed by chronological age. This makes intuitive sense because a model trained to predict age will learn to identify age-related changes, but these changes could occur more rapidly in some patients than in others. Hence, a positive AI Age Discrepancy reflects that a patient is predicted to be older than expected, which could reflect poorer biological health than is typical for their age.
\vspace{-0.7cm}

\vspace{-1cm}

\subsection{Statistical tests}
Multivariate Cox proportional-hazards regression was used to evaluate the association between AI Age Discrepancy and our two endpoints, length of hospital stay and overall survival, and the models included established covariates for each outcome. For length of stay, the covariates include surgical approach (laparoscopic/robotic vs open nephrectomy), non-nephron sparing vs nephron-sparing surgery, CCI points, tumor size, and chronological age. For overall survival, additional covariates included tumor stage (greater than or equal to 3 or less than 3), lymph node involvement, metastasis, and ordinal ISUP grade. Stage and grade were not included in hospital stay predictions as they are not known before surgical intervention. The Lifelines Python library was used for statistical analysis.

\section{Results}
\begin{figure}
\centering
\begin{subfigure}{.5\textwidth}
  \centering
  \includegraphics[width=0.9\linewidth]{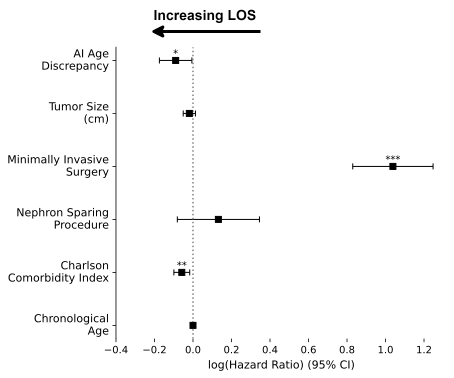}
  \caption{}
  \label{fig:sub1}
\end{subfigure}%
\begin{subfigure}{.5\textwidth}
  \centering
  \includegraphics[width=0.9\linewidth]{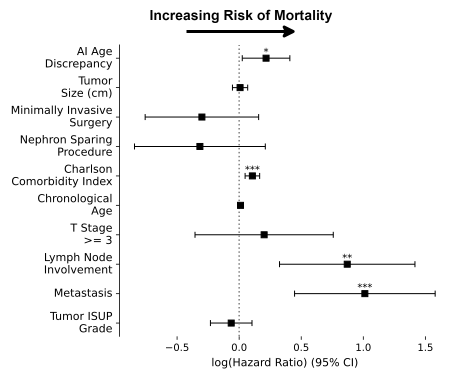}
  \caption{}
  \label{fig:sub2}
\end{subfigure}
\caption{\textbf{Forest plot analysis of (a) LOS and (b) OS.} This summarizes the results of the Cox proportional hazards regression to identify factors associated with LOS and OS following kidney cancer surgery. The log(HR) and 95\% confidence intervals are presented for each variable. (a) Lower log(HR) values indicate factors associated with a longer LOS. (b) Higher log(HR) values indicate factors associated with an increasing risk of mortality. }
\label{fig:test}
\end{figure}

We analyzed data from 590 patients who underwent surgical intervention for kidney cancer between 2010 and 2021 using Cox proportional hazards regression to identify factors associated with LOS and OS. \textit{Figure \ref{fig:test}} summarizes the Cox regression analysis of AI Age Discrepancy as a predictor of LOS and OS in a multivariate model. 

\vspace{-1cm}

\FloatBarrier

\subsection{Length of Hospital Stay}
Our analysis revealed a statistically significant association between AI Age Discrepancy and LOS. \textit{Table \ref{tab:table1}} summarizes the results of the Cox regression analysis of variables associated with LOS. The hazard ratio (HR) for AI Age Discrepancy was 0.914 (95\% CI: 0.840 to 0.994, p=0.036). Minimally invasive surgery was also associated with a significantly shorter LOS when compared to other surgical approaches with a HR of 2.825 (95\% CI: 2.293 to 3.480, p<0.001). As expected, higher Charlson Comorbidity Index (CCI) scores, which reflect a greater burden of comorbidities, were associated with a lower likelihood of early discharge (HR=0.944, 95\% CI: 0.905 to 0.983, p=0.006). The presence of a statistically significant association between length of stay and our proposed AI Age Discrepancy independent of these known predictors demonstrates the potential of our approach to provide complementary information.
\vspace{-0.6cm}
\begin{table}[h!]
\centering
\caption{\textbf{Cox regression analysis of LOS.} This summarizes the results of the Cox proportional hazards regression to identify variables associated with LOS following kidney cancer surgery.}
\begin{tabular}{c@{\hskip 0.5cm} c@{\hskip 0.5cm} c@{\hskip 0.5cm} c} 
 \hline
 Variable & Hazard Ratio & 95\% CI & p-value \\ [0.5ex] 
 \hline
 AI Age Discrepancy & 0.914 & (0.840 - 0.994) & \textbf{0.036} \\ 
 Tumor Size & 0.981 & (0.950 - 1.013) & 0.240 \\ 
 Minimally Invasive Surgery & 2.825 & (2.293 - 3.480) & \textbf{<0.001} \\
 Nephron Sparing Procedure & 1.141 & (0.922 - 1.412) & 0.225 \\
 Charlson Comorbidity Index & 0.944 & (0.905 - 0.983) & \textbf{0.006} \\
 Chronological Age & 1.000 & (0.992 - 1.008) & 0.994 \\ [1ex] 
 \hline
\end{tabular}
\label{tab:table1}
\end{table}

\vspace{-0.5cm}

\FloatBarrier

\subsection{Overall Survival}
Our analysis also revealed a statistically significant association between AI Age Discrepancy and OS. \textit{Table \ref{tab:table1}} summarizes the results of the Cox regression analysis of variables associated with OS. The HR for AI Age Discrepancy was 1.242 (95\% CI: 1.025 to 1.504, p=0.027). An HR greater than 1 indicates that a higher AI Age Discrepancy is associated with a lower likelihood of longer OS. Lymph node involvement also emerged as a significant predictor of OS (HR=2.388, 95\% CI: 1.384 to 4.120, p=0.002). Metastasis, as expected, was associated with worse prognosis (HR=2.753, 95\% CI: 1.562 to 4.848, p<0.001). Like with LOS, increased CCI points were associated with a worse prognosis (HR=1.112, 95\% CI: 1.049 to 1.180, p<0.001).  
\vspace{-0.6cm}

\begin{table}[h!]
\centering
\caption{\textbf{Cox regression analysis of OS.} This summarizes the results of the Cox proportional hazards regression to identify variables associated with OS following kidney cancer surgery.}
\begin{tabular}{c@{\hskip 0.5cm} c@{\hskip 0.5cm} c@{\hskip 0.5cm} c} 
 \hline
 Variable & Hazard Ratio & 95\% CI & p-value \\ [0.5ex] 
 \hline
 AI Age Discrepancy & 1.242 & (1.025 - 1.504) & \textbf{0.027} \\ 
 Tumor Size & 1.007 & (0.947 - 1.071) & 0.814 \\ 
 Minimally Invasive Surgery & 0.741 & (0.469 - 1.170) & 0.198 \\
 Nephron Sparing Procedure & 0.729 & (0.430 - 1.234) & 0.239 \\
 Charlson Comorbidity Index & 1.112 & (1.049 - 1.180) & \textbf{<0.001} \\
 Chronological Age & 1.011 & (0.991 - 1.031) & 0.275 \\
 T stage $\geq$ 3 & 1.223 & (0.701 - 2.134) & 0.479 \\
 Lymph Node Involvement & 2.388 & (1.384 - 4.120) & \textbf{0.002} \\
 Metastasis & 2.753 & (1.562 - 4.848) & \textbf{<0.001} \\
 Tumor ISUP Grade & 0.938 & (0.793 - 1.109) & 0.454 \\ [1ex] 
 \hline
\end{tabular}
\label{tab:table2}
\end{table}

\vspace{-0.6cm}

\FloatBarrier

\vspace{2cm}

\section{Discussion}
This study introduces AI Age Discrepancy, a novel parameter for quantifying frailty in patients undergoing surgery for kidney cancer. This approach is based on the observation that deep neural networks can learn to predict patient age with reasonable accuracy from only an abdominal pre-operative CT scan \cite{Kerber2023}. We hypothesized that patients whose AI Age Discrepancy is higher would be at a greater risk for poor postoperative outcomes following abdominal surgery like partial or radical nephrectomy. 

Our analysis reveals a statistically significant association between AI Age Discrepancy and LOS. Patients with a higher AI Age Discrepancy had a longer LOS (HR < 1). While this study is observational and causal relationships cannot be established, the findings suggest that AI Age Discrepancy could be a valuable variable to consider when making decisions regarding the surgical management of renal masses. Future prospective studies are needed to validate the clinical utility of AI Age Discrepancy for predicting postoperative outcomes. 

One possible explanation for the accuracy of our machine learning model in estimating patient age is the consideration of sarcopenia, a condition characterized by the loss of skeletal muscle mass and function. Sarcopenia is associated with aging and various chronic conditions, impacting overall health and physical performance \cite{Sabatino2021}\cite{Dutta1997}. Multiple metabolic processes and physical abilities are impacted by the loss of skeletal muscle mass, serving as a potential indicator of biological aging \cite{Wilkinson2018}. CT imaging, particularly the assessment of muscle areas such as the erector spinae at the T12 vertebra, provides a reliable measure of muscle mass and can reflect sarcopenia severity \cite{Cao2022}. 

In addition to sarcopenia, other anatomical and physiological features captured on CT imaging may contribute to the model's accuracy. For instance, the vertebral column can provide insights into degenerative changes and bone density, both of which correlate with aging as a marker of osteopenia or osteoporosis. The development of osteophytes on the vertebral column is a common age-related change and can be accurately assessed using deep learning algorithms, improving age estimation significantly \cite{Kawashita2024}. The volume of parenchymal tissue, which includes the functional parts of organs such as the kidneys, can also indicate age-related changes. Furthermore, visceral fat, which accumulates around internal organs, is known to increase with age and is associated with various metabolic diseases \cite{Cao2022}.  

We postulate that by integrating these diverse data points from CT scans, our machine-learning model can capture a comprehensive picture of a patient's biological age and health status. This multifaceted approach potentially enhances the model's predictive power for both age and associated comorbidities, ultimately improving the accuracy of post-surgical outcome predictions. The survival analysis identified established factors like lymph node involvement and metastasis as significant predictors of post-surgical survival. This aligns with existing knowledge about the association between these factors and a poorer prognosis in kidney cancer patients \cite{Campbell2021}. Importantly, our proposed AI Age Discrepancy was a statistically significant predictor independent of these factors.  

This study demonstrates the potential of deep learning and CT-derived information for clinical decision-making in kidney cancer surgery. We anticipate that the integration of AI Age Discrepancy as a predictor of frailty into clinical decision-making processes could play a significant role in improving patient outcomes, including shared-decision making regarding surgical approach as well as anticipating higher risks for complications and, thus, closer preventative monitoring of patients with higher AI Age Discrepancies. This variable might be useful for clinicians when deciding which patients to operate on and how aggressive a surgery they might tolerate.

Future research is needed to externally validate our retrospective findings, and the clinical utility of AI Age Discrepancy could be evaluated using prospective studies in which predicted age information is considered during treatment planning. Additionally, investigating the biological mechanisms underlying the association between AI Age Discrepancy and outcomes could carry significant implications once further elucidated. Finally, given the broad utility of assessing patient frailty, similar approaches may also prove useful for other surgical procedures.

\section{Conclusion}
This study presents a novel parameter, AI Age Discrepancy, as a method for quantifying the frailty and postoperative risk associated with kidney tumor surgery. Our findings reveal that a higher AI Age Discrepancy is associated with both longer hospital stays and shorter overall survival. The deep network's accuracy in estimating age likely results from its ability to capture age-related features on CT scans, such as muscle mass, bone density, and organ volume. By integrating this information, the model quantifies a more comprehensive picture of a patient's biological health, which may be useful in objectively stratifying the risk associated with invasive procedures such as partial and radical nephrectomy. This study highlights the potential for deep learning and CT-derived information to improve surgical decision-making in kidney cancer. Future research is needed to validate these findings on an external cohort, and prospective studies could be conducted to more thoroughly probe the clinical utility of considering this variable in decisions for kidney tumor care. 
\vspace{4cm}
\begin{credits}
\subsubsection{\ackname} 
\end{credits}

\bibliographystyle{plain}
\bibliography{export}

\end{document}